\documentclass[11pt]{article}

\usepackage[preprint]{acl}

\usepackage{times}
\usepackage{latexsym}

\usepackage[T1]{fontenc}

\usepackage[utf8]{inputenc}

\usepackage{microtype}

\usepackage{inconsolata}

\usepackage{graphicx}

\usepackage{amsmath}
\usepackage{amssymb}
\usepackage{booktabs}
\usepackage{listings}

\lstdefinestyle{promptstyle}{
    basicstyle=\ttfamily\footnotesize,
    breaklines=true,
    columns=fullflexible,
    keepspaces=true,
    frame=single,
    backgroundcolor=\color{gray!5},
    xleftmargin=0.5em,
    xrightmargin=0.5em,
    framexleftmargin=0.5em,
    framexrightmargin=0.5em
}

\title{Trust the Right Teacher: Quality-Aware Self-Distillation for GUI Grounding}

\author{
  Jingyuan Huang$^{1,2}$\quad
  Zuming Huang$^{2}$\quad
  Yucheng Shi$^{3}$ \\
  \textbf{Tianze Yang}$^{1}$\quad
  \textbf{Xiaoming Zhai}$^{1}$\quad
  \textbf{Wei Chu}$^{2}$\quad
  \textbf{Ninghao Liu}$^{4}$ \\
  $^{1}$University of Georgia\quad
  $^{2}$INFLY Tech\quad
  $^{3}$Tencent AI Lab \\
  $^{4}$The Hong Kong Polytechnic University
}

\begin{document}

\maketitle
\begin{abstract}
GUI grounding requires vision-language models (VLMs) to identify small target elements in high-resolution screenshots and predict precise screen coordinates. On-policy self-distillation (OPSD) is a promising post-training approach for this coordinate-sensitive task because it provides dense token-level teacher signals beyond hard coordinate labels. However, naive OPSD is not well-suited to GUI grounding: since the teacher is evaluated on student-generated prefixes, its coordinate-token signals can become unreliable once a prefix deviates from the target coordinate. We propose quality-aware self-distillation for GUI grounding, centered on direct spatial correctness verification of teacher predictions. A soft gate checks whether the teacher's top coordinate-token prediction can still be completed into the ground-truth box under the student-generated prefix and down-weights the signal when this check fails. Teacher-probability scaling then uses the teacher's confidence to further calibrate the gated supervision. A key empirical finding is that neither component alone improves overall performance, whereas combining them consistently does. This suggests that the two mechanisms play complementary roles: gating suppresses unreliable coordinate-token supervision, while probability scaling calibrates the remaining signals. Experiments across six GUI grounding benchmarks show that our method consistently improves the base model and outperforms strong baselines.
\end{abstract}

\section{Introduction}

GUI grounding is a fundamental capability for VLMs and agents that operate computers, mobile devices, and web applications
\citep{cheng2024seeclick,hong2024cogagent,gou2025uground,wu2024osatlas}.
Given a screenshot and an instruction, the model must identify the intended interface element and output its screen coordinates
\citep{cheng2024seeclick,gou2025uground,park2025rvlm}.
This task is especially challenging in high-resolution screenshots and complex GUI scenes, where target elements can be small, visually similar, and densely arranged
\citep{hong2024cogagent,li2025screenspotpro,park2025rvlm}.
Existing post-training methods provide limited supervision for this coordinate-sensitive task
\citep{park2025rvlm,zhou2025guig1,tang2025guig2}.
SFT is simple and stable, but it treats the annotated coordinate as a hard target, providing little information beyond the final answer
\citep{park2025rvlm}.
It does not exploit the teacher's uncertainty or other ``dark knowledge'' over plausible coordinate tokens
\citep{hinton2015distilling}, limiting the richness of supervision for fine-grained localization.
Reinforcement learning methods such as GRPO optimize task outcomes, but they require multiple rollouts and rely on sparse rewards
\citep{shao2024deepseekmath,zhao2026selfdistilled}.
Such outcome-only supervision is costly and provides weak guidance for fine-grained localization
\citep{tang2025guig2,park2025rvlm}.

On-policy self-distillation (OPSD) is a promising alternative to both GRPO and SFT.
By training on teacher distributions along student-generated trajectories, OPSD provides dense token-level teacher signals without requiring the large number of rollouts used by GRPO-style methods
\citep{zhao2026selfdistilled}.
In principle, such soft supervision can carry richer information than hard-label SFT, including preferences among plausible locations
\citep{hinton2015distilling,zhao2026selfdistilled}.
However, a naive instantiation of OPSD is not well suited to GUI grounding.
The effectiveness of OPSD depends critically on the quality of teacher signals.
OPSD queries the teacher on prefixes generated by the student
\citep{zhao2026selfdistilled}.
Since GUI coordinates are produced autoregressively, an incorrect student prefix can already encode a wrong spatial hypothesis.
Conditioned on such a prefix, the teacher's subsequent logits may become a plausible continuation of the wrong coordinate rather than a useful teacher signal toward the true target.
Thus, directly applying OPSD to GUI grounding can lead to unreliable coordinate-token teacher signals.

To improve the quality of teacher signals, we propose \textbf{quality-aware self-distillation} for VLM-based GUI grounding, which calibrates coordinate-token supervision by combining two complementary components: \emph{soft correctness-aware gating} and \emph{teacher-probability scaling}.
Specifically, soft correctness-aware gating exploits a special property of GUI grounding: coordinate predictions are spatially verifiable against the ground-truth bounding box
\citep{cheng2024seeclick,wu2024osatlas,tang2025guig2}.
Under the current student-generated prefix, we regard a coordinate-token teacher signal as reliable if the teacher's current coordinate-token prediction can still fall inside the ground-truth bounding box, and as unreliable if it can no longer fall inside the box.
Soft correctness-aware gating assigns the full gate value to reliable signals and down-weights unreliable ones rather than discarding them.
In addition, teacher-probability scaling uses the teacher probability of the top coordinate-token prediction as a lightweight scaling factor to further refine the gated teacher signal, assigning larger distillation weights to higher-probability teacher signals and softening lower-probability ones.
This combination makes coordinate-token supervision both correctness-aware and certainty-aware, preserving useful distributional information while reducing the negative impact of unreliable teacher signals.
The contributions of this paper are as follows:
\begin{itemize}
    \item We propose \emph{quality-aware self-distillation} for GUI grounding, using direct, prefix-aware spatial correctness verification to identify incompatible coordinate-token teacher signals and reduce their influence during training. Soft correctness-aware gating and teacher-probability scaling jointly determine the strength of teacher supervision. \item We use GUI grounding as a spatially verifiable setting to empirically study teacher signal reliability in on-policy self-distillation, especially how unreliable teacher signals should be treated during training.
    \item We conduct comprehensive experiments on six GUI grounding evaluation sets and show that our method consistently improves the base model and outperforms strong post-training baselines, with ablation studies verifying the complementary effects of gating and scaling.
\end{itemize}

\begin{figure*}[t]
    \centering
    \includegraphics[width=\textwidth]{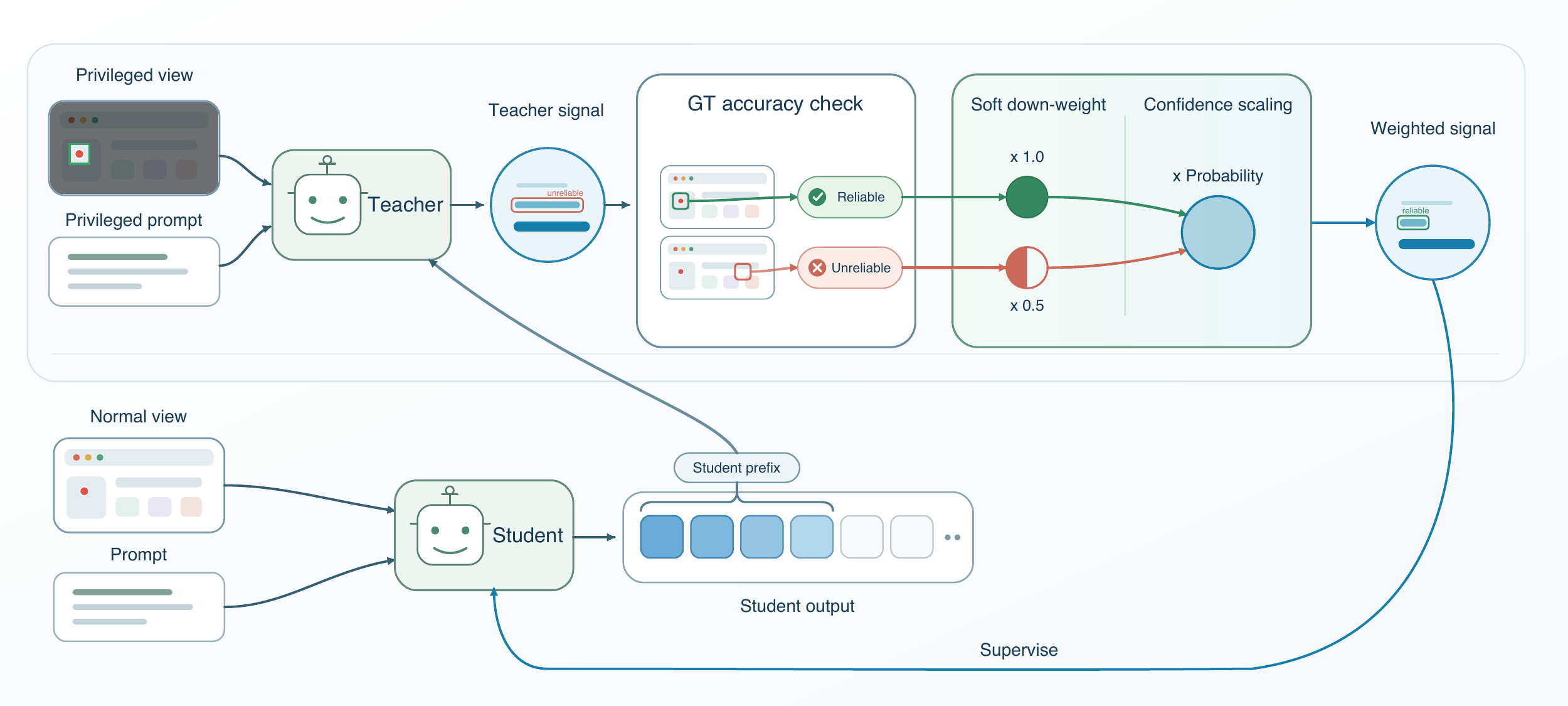}
    \caption{Overview of our proposed method. The signal acquisition process is simplified in this illustration; in practice, the signal is obtained by computing the reverse KL divergence between the probability distributions induced by the student and teacher model logits.}    \label{fig:papermethod}
\end{figure*}

\section{Related Work}

\paragraph{GUI Grounding.}
GUI grounding requires vision-language models to localize target interface elements and output precise screen coordinates given a screenshot and a natural-language instruction.
Recent work has improved GUI grounding through post-training on GUI-specific data.
Supervised fine-tuning methods train models with annotated instruction-coordinate pairs, as in SeeClick, CogAgent, UGround, OS-Atlas, and RVLM
\citep{cheng2024seeclick,hong2024cogagent,gou2025uground,wu2024osatlas,park2025rvlm}.
More recent reinforcement-learning-based methods further optimize GUI grounding with verifiable reward signals~\citep{yang2026trontargetedruleverifiableonline}, including GUI-G1 and GUI-G2
\citep{zhou2025guig1,tang2025guig2}.
Some works further design distance-aware or continuous spatial rewards for GRPO-style GUI grounding, providing denser feedback according to how close the predicted coordinate is to the target region
\citep{shi2026trustworthyguiagentssurvey,zeng2026fdcground,tang2025guig2,zhao2025guicursor}.
\paragraph{Self-Distillation and Teacher-Signal Reliability.}
OPSD trains the student on trajectories sampled from its own policy, while a teacher model provides token-level teacher signals on the same student-generated prefixes
\citep{zhao2026selfdistilled,yuan2026visionopd}.
Prior On-Policy Distillation (OPD)/OPSD analyses point out that teacher signals are not uniformly reliable
\citep{zhu2026manyfaces,zheng2026scope,ke2026respecting}.
In particular, when the teacher is conditioned on student-generated prefixes, these prefixes may be imperfect or distributionally mismatched, causing the teacher signal to become noisy or less informative
\citep{zhu2026manyfaces,zheng2026scope}.
Other work further observes that directly exposing the full correct answer can make the privileged teacher signal overly sharp or near-deterministic, weakening the benefit of soft distillation
\citep{tan2026paint,zhang2026guisd}.

\paragraph{Methods to Improve Teacher-Signal Reliability.}To mitigate these problems, several reliability-aware OPD/OPSD methods improve teacher signals through proxy-based weighting or privileged teacher inputs.
For example, entropy-based self-distillation methods use teacher uncertainty to adjust token-level update weights
\citep{ke2026respecting,zhang2026guisd};
perplexity-based OPD methods down-weight teacher guidance that appears unreliable
\citep{zheng2026scope};
and methods that combine self-distillation with verifiable feedback use outcome correctness to anchor update directions or route supervision paths
\citep{yang2026selfdistilledrlvr,zheng2026scope}.
In vision-language settings, Vision-OPD constructs a crop-conditioned teacher to provide more focused teacher signals for a full-image student
\citep{yuan2026visionopd}.
GUI-SD further adapts this idea to GUI grounding by constructing a visually enriched privileged teacher input with layout-preserving regional masking, while using coordinate-token weighting and entropy-based scaling to improve GUI self-distillation
\citep{zhang2026guisd}.
These works establish an important point: naive self-distillation should not treat all teacher signals as equally trustworthy.
However, existing criteria still mainly rely on indirect proxies to improve teacher signals' quality.
Indirect proxies such as entropy, teacher probability, or perplexity may correlate with signal reliability on average, but they do not provide a direct guarantee that the selected or emphasized signals are actually reliable.

GUI grounding provides a natural opportunity to bridge this gap, because coordinate predictions are spatially verifiable.
Our work leverages this structure and uses the ground-truth box as a direct training-time reliability criterion for coordinate-token teacher signals.
Instead of blindly imitating unreliable teacher signals, our method softly down-weights and further scales their loss contribution before distillation.

\section{Methodology}
\label{sec:method}

\subsection{Privileged Information Construction}

For each training example, we are given a GUI instruction $x$, a screenshot $I$, and a ground-truth bounding box $B$. The student is conditioned on the original input $(x, I)$. During training, following GUI-SD's visually privileged input design, we additionally construct a privileged teacher input $(x^+, I^+)$, where $I^+$ is obtained by \emph{layout-preserving regional masking}: the target region is preserved and visually highlighted, while task-irrelevant regions are suppressed~\citep{zhang2026guisd}. The teacher-only prompt $x^+$ indicates that the answer lies in the highlighted region and asks the teacher to answer the question. At inference time, only the original input $(x, I)$ is available.

\subsection{Constructing Quality-aware Teacher Signals}

The student samples an on-policy response $y$, and both student and teacher distributions are evaluated on the same student-generated prefix:
\begin{equation}
\label{eq:student-teacher-distributions}
\begin{aligned}
y &\sim \pi_\theta(\cdot \mid x, I),\\
P_S^t &= \pi_\theta(\cdot \mid x, I, y_{<t}),\\
P_T^t &= \operatorname{sg}\!\left[\pi_\theta(\cdot \mid x^+, I^+, y_{<t})\right].
\end{aligned}
\end{equation}
Here, $\operatorname{sg}[\cdot]$ denotes stop-gradient, so the privileged teacher distribution is used only as a teacher signal.

The core of our method is to combine soft correctness-aware gating with teacher-probability scaling for coordinate-token supervision. This mechanism is applied only to coordinate tokens, i.e., coordinate digit tokens in the model response. Other response tokens, such as formatting tokens or non-coordinate text tokens, are distilled normally. This design focuses on coordinate-token supervision, which directly determines GUI grounding accuracy.

\paragraph{Soft correctness-aware gating.}

For a coordinate-token position $t$, we first apply softmax to the privileged teacher's logits produced under the current student-generated prefix, obtaining the teacher distribution $P_T^t$. Let $\mathcal{D}$ denote the set of coordinate digit tokens. We define $d_t^\star$ as the coordinate digit token with the highest teacher probability:
\begin{equation}
\label{eq:top-coordinate-token}
d_t^\star = \arg\max_{d \in \mathcal{D}} P_T^t(d).
\end{equation}

The binary compatibility indicator $h_t$ then verifies whether this top coordinate-token prediction remains feasible for the corresponding coordinate axis under the current student-generated prefix. For a coordinate-token position $t$, let $a(t) \in \{\mathrm{x}, \mathrm{y}\}$ denote its axis, and let $B_{a(t)}$ denote the interval of the ground-truth bounding box $B$ on axis $a(t)$. We append $d_t^\star$ to the current axis-specific prefix induced by $y_{<t}$, and check whether the remaining digits can still be completed into a valid coordinate value within $B_{a(t)}$. Thus, $\mathrm{x}$-coordinate tokens are checked only against the $\mathrm{x}$-axis interval of $B$, and $\mathrm{y}$-coordinate tokens only against the $\mathrm{y}$-axis interval. We define
\begin{equation}
\label{eq:compatibility-indicator}
h_t =
\begin{cases}
1, & \text{if such a completion exists,}\\
0, & \text{otherwise.}
\end{cases}
\end{equation}
This prefix-aware binary indicator identifies whether the teacher's strongest coordinate-token prediction is compatible with the target region under the current student-generated prefix.

Instead of using hard correctness-aware gating, which would discard failed-gate coordinate-token signals entirely, we convert the binary compatibility indicator into a soft correctness-aware gate:
\begin{equation}
\label{eq:soft-gate}
g_t = \alpha + (1-\alpha)h_t.
\end{equation}
Thus, compatible coordinate-token predictions receive gate value $g_t=1$, while incompatible coordinate-token predictions receive gate value $g_t=\alpha$ rather than being discarded. In our main method, we set $\alpha=0.5$, so failed-gate coordinate-token signals are down-weighted by half. This soft correctness-aware gating strategy preserves potentially useful teacher signals while reducing the influence of unreliable coordinate-token supervision.

\paragraph{Teacher-probability scaling.}
However, correctness-aware gating alone is still insufficient. The gate provides a prefix-aware judgment of spatial compatibility, but it does not measure the teacher's uncertainty. Two teacher predictions may both pass the gate, while their distributions can have very different quality: a higher teacher probability usually reflects a clearer preference, whereas a lower teacher probability may be closer to a decision boundary and more likely to be affected by visual clutter, occlusion, or similar distractors \citep{hendrycks2017baseline,guo2017calibration,zheng2026scope,ke2026respecting}. Therefore, for coordinate-token positions, we further scale the distillation strength by the teacher probability. We define
\begin{equation}
\label{eq:teacher-probability}
p_t = P_T^t(d_t^\star),
\end{equation}
where $p_t$ is the probability assigned by the privileged teacher to its top-1 coordinate-token prediction. A larger $p_t$ indicates that the teacher assigns higher probability to the current coordinate token, so the corresponding distillation term should contribute more strongly; a smaller $p_t$ indicates higher uncertainty, so the teacher signal is softened even when it is spatially compatible with the target box.

\subsection{Weighted Reverse-KL Objective}

Combining the coordinate-token indicator, the soft correctness-aware gate, and the teacher-probability scaling term, we define the token-level distillation weight as
\begin{equation}
\label{eq:token-weight}
w_t = (1-r_t) + r_t\, g_t\, \lambda\, p_t.
\end{equation}
Here, $r_t$ indicates whether position $t$ is a coordinate token. If $t$ is not a coordinate token, then $r_t=0$ and $w_t=1$, so the token is distilled normally. If $t$ is a coordinate token, then $r_t=1$, and the loss contribution is controlled by the soft correctness-aware gate $g_t$ and the teacher probability $p_t$. The scaling coefficient $\lambda$, a fixed scalar, further calibrates the overall contribution of coordinate-token supervision. Since reliability-based weighting can reduce the aggregate loss mass assigned to coordinate tokens, $\lambda$ prevents these decisive tokens from becoming under-emphasized in the training objective. $\lambda$ is set as 3 in our main experiment.

Equivalently, for coordinate and non-coordinate tokens, Eq.~\eqref{eq:token-weight} gives
\begin{equation}
\label{eq:weight-behaviors}
w_t =
\begin{cases}
1, & r_t=0,\\
\lambda p_t, & r_t=1 \text{ and } h_t=1,\\
\alpha \lambda p_t, & r_t=1 \text{ and } h_t=0.
\end{cases}
\end{equation}
Finally, we train the student with the resulting weighted reverse-KL objective over response tokens:
\begin{equation}
\label{eq:ours-loss}
\begin{aligned}
\mathcal{L}_{\text{ours}}(\theta)
&= \mathbb{E}_{y \sim \pi_\theta(\cdot \mid x,I)}\!\left[
\frac{1}{|\mathcal{R}(y)|}
\right.\\[-0.25em]
&\qquad\left.
\sum_{t \in \mathcal{R}(y)} w_t\, D_{\mathrm{KL}}\!\left(P_S^t \,\|\, P_T^t\right)
\right].
\end{aligned}
\end{equation}
Here, $\mathcal{R}(y)$ denotes the set of response-token positions. Prompt tokens are excluded from the loss. The expectation is over the on-policy student response $y$ sampled from the student policy, as defined in Eq.~\eqref{eq:student-teacher-distributions}.

Eq.~\eqref{eq:token-weight} yields three behaviors: (i) ordinary response tokens are distilled with weight $1$; (ii) compatible coordinate tokens are distilled with weight $\lambda p_t$; and (iii) incompatible coordinate tokens are down-weighted to $\alpha \lambda p_t$. In our main method, we set $\alpha=0.5$ and $\lambda=3$.  The soft correctness-aware gating and teacher-probability scaling mechanism only modulates the loss contribution of each token-level KL term, thereby reducing blind imitation of unreliable coordinate-token supervision while still preserving useful teacher signals. 

\section{Experiments}
\label{sec:experiments}

\subsection{Experimental Setup}

Our primary experiments use Qwen3.5-9B as the backbone model. We additionally evaluate our method on Qwen3-VL-4B-Instruct to examine whether the proposed supervision weighting also benefits another backbone. The resulting Qwen3.5-9B checkpoint is released under the Apache License 2.0, consistent with the license of its base model. The training data follows the GUI-SD's data construction, which is built based on ScaleCUA~\citep{zhang2026guisd,liu2025scalecua}. We report results on six GUI grounding benchmarks: ScreenSpot-Pro (SSP) \citep{li2025screenspotpro}, ScreenSpot-v2 \citep{wu2024osatlas}, UI-Vision Element Grounding (UIEG) \citep{nayak2025uivision}, OSWorld-G, OSWorld-G-R \citep{xie2025osworldg}, and MMBench-GUI L2 Element Grounding (MMG) \citep{wang2026mmbenchgui}. UIEG and MMG are GUI grounding subsets of their corresponding benchmarks \citep{nayak2025uivision,wang2026mmbenchgui}.

\subsection{Main Results}

Table~\ref{tab:main_results} shows that our method achieves the best results across six benchmarks among all compared methods. Our method reaches 72.23 macro-average accuracy, outperforming the strongest baseline, the GUI-SD baseline, by 2.16 points. These results demonstrate that improving the quality and reliability of the teacher signal is an effective direction for GUI grounding OPSD.

\begin{table*}[t]
  \centering
  \small
  \setlength{\tabcolsep}{3pt}
    \begin{tabular}{lccccccc}
      \toprule
      Method
      & SSP
      & ScreenSpot-v2
      & UIEG
      & OSWorld-G
      & OSWorld-G-R
      & MMG
      & Avg \\
      \midrule
      Qwen3.5-9B
      & 63.00 & 91.75 & 26.92 & 61.35 & 67.73 & 80.41 & 65.19 \\
      GRPO
      & 63.50 & 91.90 & 28.36 & 61.52 & 68.26 & 81.64 & 65.86 \\
      SFT
      & 65.15 & 93.32 & 33.33 & 62.06 & 71.10 & 83.59 & 68.09 \\
      Naive-OPSD
      & 66.41 & 93.87 & 41.87 & 59.75 & 65.60 & 85.98 & 68.91 \\
      GUI-SD
      & 64.71 & 94.97 & 42.29 & 63.65 & 70.04 & 84.76 & 70.07 \\
      Ours
      & \textbf{68.37} & \textbf{95.68} & \textbf{43.24} & \textbf{66.49} & \textbf{72.34} & \textbf{87.26} & \textbf{72.23} \\
      \bottomrule
    \end{tabular}
  \caption{
    Main results on six GUI grounding evaluation sets.
    Avg denotes macro-average accuracy, computed as the arithmetic mean over the six evaluation sets.
  }
  \label{tab:main_results}
\end{table*}

\begin{table*}[t]
  \centering
  \small
  \setlength{\tabcolsep}{3pt}
  \begin{tabular}{lccccccc}
    \toprule
    Method & SSP & ScreenSpot-v2 & UIEG & OSWorld-G & OSWorld-G-R & MMG & Avg \\
    \midrule
    Base model & 59.33 & 92.30 & 24.18 & 59.04 & 67.73 & 83.44 & 64.34 \\
    Vision-PV-Only & 61.42 & 93.35 & 29.38 & 61.52 & 69.15 & 84.74 & 66.59 \\
    Ours & \textbf{62.80} & \textbf{93.79} & \textbf{29.88} & \textbf{62.40} & \textbf{69.86} & \textbf{85.02} & \textbf{67.29} \\
    \bottomrule
  \end{tabular}
  \caption{Results on Qwen3-VL-4B-Instruct across six GUI grounding benchmarks.}
  \label{tab:qwen3vl4b_results}
\end{table*}

Compared with the GUI-SD baseline, the key difference lies in how the teacher signal is weighted. GUI-SD strengthens coordinate-token supervision through digit-position weighting and entropy-based scaling~\citep{zhang2026guisd}. However, these weights are not explicitly correctness-aware: a teacher signal can still be amplified even when it is inconsistent with the target coordinate or potentially harmful to the student. Our method instead calibrates the teacher signal according to signal reliability. By reducing the influence of unreliable teacher signals and emphasizing more reliable coordinate-token supervision, our method provides a higher-quality teacher signal for GUI grounding. Our ablation studies will further elaborate this.

Our method improves macro-average accuracy over SFT by 4.14 percentage points and over the GRPO checkpoint at step 62 reported in Table~\ref{tab:main_results} by 6.37 percentage points. Appendix~\ref{app:training_details} additionally reports a full-epoch GRPO run, whose best evaluated checkpoint reaches 68.30\%, compared with 72.23\% for our method. Compared with SFT, which mainly learns from hard target labels under teacher forcing, our self-distillation objective leverages teacher logits as soft supervision. Such soft targets contain richer ``dark knowledge'' beyond one-hot labels and can provide more informative training signals for the student~\citep{hinton2015distilling}. Moreover, since our method supervises the student on its own generated prefixes, it better matches the autoregressive inference process and helps mitigate the exposure-bias problem, where training on ground-truth prefixes but testing on model-generated prefixes may lead to error accumulation~\citep{bengio2015scheduled,zhao2026selfdistilled,zhang2026guisd}. Compared with GRPO, our method provides dense token-level supervision from the teacher distribution, whereas GRPO mainly relies on sparse outcome-level reward feedback~\citep{shao2024deepseekmath,zhang2026guisd}. This dense supervision in our method is particularly beneficial for GUI grounding, where accurate coordinate prediction requires fine-grained token-level learning signals.

\paragraph{Results on a second backbone.}
We further compare our method with Vision-PV-Only, the unweighted visual-privilege self-distillation baseline, on Qwen3-VL-4B-Instruct. Both runs share the same initialization, training data, privileged teacher inputs, and training and evaluation configurations, differing only in supervision weighting. As shown in Table~\ref{tab:qwen3vl4b_results}, at the designated final checkpoint after 62 training steps, our method improves macro-average accuracy from 66.59\% to 67.29\%, a gain of 0.70 percentage points, with improvements on all six benchmarks. These results provide additional evidence that quality-aware teacher weighting benefits GUI grounding beyond Qwen3.5-9B.

\subsection{Ablation Studies}

\begin{table*}[t]
  \centering
  \small
  \setlength{\tabcolsep}{3pt}
    \begin{tabular}{lccccccccc}
      \toprule
      Method
      & Gating
      & Scaling
      & SSP
      & ScreenSpot-v2
      & UIEG
      & OSWorld-G
      & OSWorld-G-R
      & MMG
      & Avg \\
      \midrule
      Vision-PV-Only
      & None
      & None
      & 67.49 & 94.10 & 41.74 & 63.65 & 70.04 & 85.57 & 70.43 \\
      +Soft Gating
      & Soft
      & None
      & 67.11 & 94.50 & 41.58 & 61.88 & 69.68 & 85.05 & 69.97 \\
      +Probability Scaling
      & None
      & $3p_t$
      & 67.24 & 94.18 & 41.19 & 63.65 & 70.74 & 84.16 & 70.19 \\
      +Gating \& Scaling (Ours)
      & Soft
      & $3p_t$
      & \textbf{68.37} & \textbf{95.68} & \textbf{43.24} & \textbf{66.49} & \textbf{72.34} & \textbf{87.26} & \textbf{72.23} \\
      \bottomrule
    \end{tabular}
  \caption{
    High-level component analysis.
  }
  \label{tab:ablation_components}
\end{table*}

\begin{table*}[t]
  \centering
  \small
  \setlength{\tabcolsep}{3pt}
    \begin{tabular}{lccccccccc}
      \toprule
      Method
      & Gating
      & Scaling
      & SSP
      & \shortstack{ScreenSpot-v2}
      & UIEG
      & OSWorld-G
      & \shortstack{OSWorld-G-R}
      & MMG
      & Avg \\
      \midrule
      \shortstack[l]{Probability Scaling}
      & None
      & $3p_t$
      & 67.24 & 94.18 & 41.19 & 63.65 & 70.74 & 84.16 & 70.19 \\
      \shortstack[l]{+Soft Gating (Ours)}
      & Soft, $\alpha=0.5$
      & $3p_t$
      & \textbf{68.37} & \textbf{95.68} & \textbf{43.24} & \textbf{66.49} & 72.34 & \textbf{87.26} & \textbf{72.23} \\
      \shortstack[l]{+Hard Gating}
      & Hard, $\alpha=0$
      & $3p_t$
      & 67.87 & 95.05 & 43.11 & 64.54 & \textbf{72.52} & 85.67 & 71.46 \\
      \bottomrule
    \end{tabular}
  \caption{
    Effect of gating strength.
  }
  \label{tab:ablation_gating}
\end{table*}

\begin{table*}[t]
  \centering
  \small
  \setlength{\tabcolsep}{1.5pt}
    \begin{tabular}{lccccccccc}
      \toprule
      Method
      & Gating
      & \shortstack{Coordinate-Token\\Weight}
      & SSP
      & \shortstack{ScreenSpot-\\v2}
      & UIEG
      & OSWorld-G
      & \shortstack{OSWorld-\\G-R}
      & MMG
      & Avg \\
      \midrule
      Fixed $\lambda=3$
      & Soft, $\alpha=0.5$
      & $3(Pass) / 1.5(Fail)$
      & 67.74 & 94.73 & 41.03 & 65.43 & 71.99 & 85.81 & 71.12 \\
      \shortstack[l]{+ Probability scaling\\(Ours)}
      & Soft, $\alpha=0.5$
      & $3p_t(Pass) / 1.5p_t(Fail)$
      & \textbf{68.37} & \textbf{95.68} & \textbf{43.24} & \textbf{66.49} & \textbf{72.34} & \textbf{87.26} & \textbf{72.23} \\
      \bottomrule
    \end{tabular}
  \caption{
    Effect of teacher-probability scaling.
  }
  \label{tab:ablation_prob_scaling}
\end{table*}

\begin{table*}[t]
  \centering
  \small
  \setlength{\tabcolsep}{3pt}
    \begin{tabular}{lccccccc}
      \toprule
      $\lambda$
      & SSP
      & ScreenSpot-v2
      & UIEG
      & OSWorld-G
      & OSWorld-G-R
      & MMG
      & Avg \\
      \midrule
      1
      & 67.49 & 94.97 & 42.31 & 64.72 & 71.45 & 86.25 & 71.20 \\
      2
      & 67.68 & 95.05 & 42.07 & 65.25 & 71.99 & 85.87 & 71.32 \\
      3 (Ours)
      & 68.37 & \textbf{95.68} & \textbf{43.24} & \textbf{66.49} & \textbf{72.34} & \textbf{87.26} & \textbf{72.23} \\
      4
      & \textbf{69.07} & 95.52 & 42.76 & 65.78 & 71.81 & 85.88 & 71.80 \\
      \bottomrule
    \end{tabular}
  \caption{
    Effect of the scaling coefficient.
  }
  \label{tab:ablation_scaling_coeff}
\end{table*}

We organize our analysis around four research questions concerning the complementarity of correctness gating and probability scaling, the treatment of spatially incompatible teacher signals, the relationship between teacher probability and supervision quality, and the overall strength of coordinate-token supervision.

\paragraph{RQ1: How do correctness gating and teacher-probability scaling complement each other?}

Table~\ref{tab:ablation_components} compares the individual and combined effects of the two components against Vision-PV-Only, an unweighted self-distillation baseline with the same visual privileged information. The baseline achieves 70.43\% macro-average accuracy. Adding correctness gating alone reduces the average to 69.97\%, while adding teacher-probability scaling alone yields 70.19\%. Combining both components improves the average to 72.23\%, a gain of 1.80 percentage points over the baseline. These results show that the combined design is effective even though neither component alone improves the overall accuracy.

\textbf{The two components address different aspects of teacher supervision: correctness gating directly checks spatial validity, while probability scaling adjusts the strength of the teacher's guidance.} Gating alone reduces the contribution of spatially incompatible predictions, but may also weaken supervision from teacher distributions that still contain useful information. Probability scaling alone can strengthen high-confidence predictions without checking whether they are spatially correct, potentially amplifying unreliable supervision.

\textbf{Their combination couples correctness verification with selective reinforcement of teacher signals.} Under the same teacher probability, a prediction that fails the spatial check receives half the weight of one that passes. Within each gate group, higher-probability predictions receive greater weight. Together with the overall scaling coefficient, this design reduces the relative emphasis on spatially incompatible signals while maintaining stronger supervision from compatible, high-confidence predictions. The component ablations support this complementary use of correctness and probability when weighting teacher supervision.

\paragraph{RQ2: How should spatially incompatible teacher signals be weighted?}

Table~\ref{tab:ablation_gating} compares different treatments of spatially incompatible teacher signals while keeping teacher-probability scaling fixed at $3p_t$. Without correctness gating, the model achieves 70.19\% macro-average accuracy. Hard gating, which removes the coordinate-token distillation terms at failed-gate positions, improves the average to 71.46\%. Soft gating achieves the best result at 72.23\%. \textbf{Both gating variants improve over probability scaling alone, supporting the value of directly checking the spatial correctness of teacher predictions.}

Soft gating further outperforms hard gating by 0.77 percentage points. A failed spatial check identifies a teacher prediction that cannot be completed into the target region under the current student prefix, but does not establish that its entire distribution is devoid of useful information. Completely removing the corresponding distillation term may therefore discard potentially useful supervision. \textbf{Soft gating provides a better balance in our experiments by reducing the contribution of these signals without eliminating it.}

\paragraph{RQ3: Does teacher probability reflect supervision quality?}

\textbf{We examine this question through both a direct analysis of teacher predictions and an ablation of probability-based weighting.} We analyze 565 coordinate-digit predictions from 100 training examples. A digit is considered spatially correct if, when appended to the student's current-axis prefix, it still permits a coordinate completion within the target interval. Teacher probability is positively correlated with this correctness measure ($r=0.301$); correct and incorrect predictions have mean probabilities of 0.708 and 0.298, respectively. These results indicate that teacher probability provides useful information about the spatial correctness of teacher predictions, supporting its use as a relative indicator of supervision quality.

\textbf{The ablation further shows that using this information improves distillation.} Table~\ref{tab:ablation_prob_scaling} keeps soft correctness-aware gating and the scaling coefficient $\lambda=3$ unchanged, and varies only whether teacher probability is included in the weight. Removing the probability term reduces macro-average accuracy from 72.23\% to 71.12\%, a decrease of 1.11 percentage points. Thus, even after direct correctness verification is applied, teacher probability remains useful for adjusting supervision strength. The correlation analysis and the controlled ablation together support using probability to further weight teacher signals within the correctness-aware framework.

\paragraph{RQ4: How does the strength of coordinate-token supervision affect performance?}

Table~\ref{tab:ablation_scaling_coeff} varies the scaling coefficient $\lambda$ while keeping correctness gating and teacher-probability weighting unchanged. Macro-average accuracy is 71.20\%, 71.32\%, 72.23\%, and 71.80\% for $\lambda=1,2,3,$ and $4$, respectively. \textbf{The best average among the tested values is obtained at $\lambda=3$, showing that the overall strength of coordinate-token supervision matters alongside its quality-aware weighting.}

Correctness gating and probability weighting reduce the contribution of incompatible or low-probability predictions. The scaling coefficient controls the overall emphasis placed on the remaining coordinate-token supervision. Increasing $\lambda$ from 1 to 3 improves the average accuracy, whereas a further increase to 4 lowers it despite improving SSP. These results support a moderate scaling coefficient that maintains sufficient coordinate-token supervision without placing excessive weight on it.

\section{Discussion}

GUI grounding provides a concrete setting for examining how teacher signals should be used in on-policy distillation when their quality can be explicitly verified. Unlike general token prediction tasks, GUI grounding has a spatially checkable structure: under a given decoding prefix, a coordinate-token prediction can be tested by whether it remains possible to complete it into the ground-truth target region. This allows teacher supervision to be assessed not only through indirect proxies such as confidence or uncertainty, but also through its compatibility with the target constraint.

Under this view, unreliable teacher signals should not be treated in a purely binary manner. A signal that is incompatible with the ground-truth region should not be imitated as strongly as a compatible one, since doing so may reinforce incorrect spatial predictions. However, such signals are not necessarily devoid of useful information; they may still reflect local preferences or distributional structure learned by the teacher. Therefore, our study explores a soft way of using teacher supervision under verifiable reliability: target compatibility is used to adjust the trust placed in the teacher signal, while probability calibration controls the strength of the supervision. This provides an initial attempt to make on-policy distillation more reliability-aware in GUI grounding, where imperfect teacher signals are weakened and reshaped rather than simply discarded.

\section{Conclusion}

We presented quality-aware self-distillation for GUI grounding, aiming to improve the reliability of coordinate-token teacher signals in on-policy self-distillation. Our method uses soft correctness-aware gating to down-weight teacher predictions that are incompatible with the target region under the student-generated prefix, and further applies teacher-probability scaling to refine the strength of coordinate-token supervision. Experiments on six GUI grounding benchmarks show that the proposed method consistently improves the base model and outperforms strong post-training baselines, including SFT, GRPO, naive OPSD, and GUI-SD. These results highlight spatial verifiability as an effective signal for improving teacher-signal reliability in GUI grounding self-distillation.

\section{Limitations}

Our study leaves several natural directions for future extension. First, the current formulation assumes that a task-specific correctness signal can be directly verified during training, as in GUI grounding with ground-truth target boxes. For tasks where correctness is not directly observable, the same quality-aware principle may require reliable proxy indicators to assess teacher-signal quality. Second, the current study adopts a reasoning-free action format and does not evaluate reasoning-enabled OPD/OPSD. In reasoning-enabled settings, correctness-aware teacher calibration may be more complex because intermediate reasoning tokens are not directly verifiable in the same way as structured actions. Extending the proposed approach to reasoning-enabled distillation therefore remains an important direction for future work. Third, our experiments focus on single-step grounding and do not cover multi-turn agentic computer-use tasks, where long-horizon interactions and error accumulation may introduce additional challenges. The effectiveness of our method in such settings therefore remains to be evaluated.
\bibliography{references}

@misc{cheng2024seeclick,
      title={SeeClick: Harnessing GUI Grounding for Advanced Visual GUI Agents}, 
      author={Kanzhi Cheng and Qiushi Sun and Yougang Chu and Fangzhi Xu and Yantao Li and Jianbing Zhang and Zhiyong Wu},
      year={2024},
      eprint={2401.10935},
      archivePrefix={arXiv},
      primaryClass={cs.HC},
      url={https://arxiv.org/abs/2401.10935}, 
}

@misc{hong2024cogagent,
      title={CogAgent: A Visual Language Model for GUI Agents}, 
      author={Wenyi Hong and Weihan Wang and Qingsong Lv and Jiazheng Xu and Wenmeng Yu and Junhui Ji and Yan Wang and Zihan Wang and Yuxuan Zhang and Juanzi Li and Bin Xu and Yuxiao Dong and Ming Ding and Jie Tang},
      year={2024},
      eprint={2312.08914},
      archivePrefix={arXiv},
      primaryClass={cs.CV},
      url={https://arxiv.org/abs/2312.08914}, 
}

@misc{gou2025uground,
      title={Navigating the Digital World as Humans Do: Universal Visual Grounding for GUI Agents}, 
      author={Boyu Gou and Ruohan Wang and Boyuan Zheng and Yanan Xie and Cheng Chang and Yiheng Shu and Huan Sun and Yu Su},
      year={2025},
      eprint={2410.05243},
      archivePrefix={arXiv},
      primaryClass={cs.AI},
      url={https://arxiv.org/abs/2410.05243}, 
}

@misc{wu2024osatlas,
      title={OS-ATLAS: A Foundation Action Model for Generalist GUI Agents}, 
      author={Zhiyong Wu and Zhenyu Wu and Fangzhi Xu and Yian Wang and Qiushi Sun and Chengyou Jia and Kanzhi Cheng and Zichen Ding and Liheng Chen and Paul Pu Liang and Yu Qiao},
      year={2024},
      eprint={2410.23218},
      archivePrefix={arXiv},
      primaryClass={cs.CL},
      url={https://arxiv.org/abs/2410.23218}, 
}

@misc{park2025rvlm,
      title={R-VLM: Region-Aware Vision Language Model for Precise GUI Grounding}, 
      author={Joonhyung Park and Peng Tang and Sagnik Das and Srikar Appalaraju and Kunwar Yashraj Singh and R. Manmatha and Shabnam Ghadar},
      year={2025},
      eprint={2507.05673},
      archivePrefix={arXiv},
      primaryClass={cs.CV},
      url={https://arxiv.org/abs/2507.05673}, 
}

@misc{li2025screenspotpro,
      title={ScreenSpot-Pro: GUI Grounding for Professional High-Resolution Computer Use}, 
      author={Kaixin Li and Ziyang Meng and Hongzhan Lin and Ziyang Luo and Yuchen Tian and Jing Ma and Zhiyong Huang and Tat-Seng Chua},
      year={2025},
      eprint={2504.07981},
      archivePrefix={arXiv},
      primaryClass={cs.CV},
      url={https://arxiv.org/abs/2504.07981}, 
}

@misc{nayak2025uivision,
      title={UI-Vision: A Desktop-centric GUI Benchmark for Visual Perception and Interaction}, 
      author={Shravan Nayak and Xiangru Jian and Kevin Qinghong Lin and Juan A. Rodriguez and Montek Kalsi and Rabiul Awal and Nicolas Chapados and M. Tamer Özsu and Aishwarya Agrawal and David Vazquez and Christopher Pal and Perouz Taslakian and Spandana Gella and Sai Rajeswar},
      year={2025},
      eprint={2503.15661},
      archivePrefix={arXiv},
      primaryClass={cs.CV},
      url={https://arxiv.org/abs/2503.15661}, 
}

@misc{xie2025osworldg,
      title={Scaling Computer-Use Grounding via User Interface Decomposition and Synthesis}, 
      author={Tianbao Xie and Jiaqi Deng and Xiaochuan Li and Junlin Yang and Haoyuan Wu and Jixuan Chen and Wenjing Hu and Xinyuan Wang and Yuhui Xu and Zekun Wang and Yiheng Xu and Junli Wang and Doyen Sahoo and Tao Yu and Caiming Xiong},
      year={2025},
      eprint={2505.13227},
      archivePrefix={arXiv},
      primaryClass={cs.AI},
      url={https://arxiv.org/abs/2505.13227}, 
}

@misc{wang2026mmbenchgui,
      title={MMBench-GUI: Hierarchical Multi-Platform Evaluation Framework for GUI Agents}, 
      author={Xuehui Wang and Zhenyu Wu and JingJing Xie and Zichen Ding and Bowen Yang and Zehao Li and Zhaoyang Liu and Qingyun Li and Xuan Dong and Zhe Chen and Weiyun Wang and Xiangyu Zhao and Jixuan Chen and Haodong Duan and Tianbao Xie and Chenyu Yang and Shiqian Su and Yue Yu and Yuan Huang and Yiqian Liu and Xiao Zhang and Yanting Zhang and Xiangyu Yue and Weijie Su and Xizhou Zhu and Wei Shen and Jifeng Dai and Wenhai Wang},
      year={2025},
      eprint={2507.19478},
      archivePrefix={arXiv},
      primaryClass={cs.CV},
      url={https://arxiv.org/abs/2507.19478}, 
}

@misc{hinton2015distilling,
      title={Distilling the Knowledge in a Neural Network}, 
      author={Geoffrey Hinton and Oriol Vinyals and Jeff Dean},
      year={2015},
      eprint={1503.02531},
      archivePrefix={arXiv},
      primaryClass={stat.ML},
      url={https://arxiv.org/abs/1503.02531}, 
}

@misc{shao2024deepseekmath,
      title={DeepSeekMath: Pushing the Limits of Mathematical Reasoning in Open Language Models}, 
      author={Zhihong Shao and Peiyi Wang and Qihao Zhu and Runxin Xu and Junxiao Song and Xiao Bi and Haowei Zhang and Mingchuan Zhang and Y. K. Li and Y. Wu and Daya Guo},
      year={2024},
      eprint={2402.03300},
      archivePrefix={arXiv},
      primaryClass={cs.CL},
      url={https://arxiv.org/abs/2402.03300}, 
}

@misc{zhou2025guig1,
      title={GUI-G1: Understanding R1-Zero-Like Training for Visual Grounding in GUI Agents}, 
      author={Yuqi Zhou and Sunhao Dai and Shuai Wang and Kaiwen Zhou and Qinglin Jia and Jun Xu},
      year={2025},
      eprint={2505.15810},
      archivePrefix={arXiv},
      primaryClass={cs.CL},
      url={https://arxiv.org/abs/2505.15810}, 
}

@misc{tang2025guig2,
      title={GUI-G$^2$: Gaussian Reward Modeling for GUI Grounding}, 
      author={Fei Tang and Zhangxuan Gu and Zhengxi Lu and Xuyang Liu and Shuheng Shen and Changhua Meng and Wen Wang and Wenqi Zhang and Yongliang Shen and Weiming Lu and Jun Xiao and Yueting Zhuang},
      year={2025},
      eprint={2507.15846},
      archivePrefix={arXiv},
      primaryClass={cs.LG},
      url={https://arxiv.org/abs/2507.15846}, 
}

@misc{zhao2026selfdistilled,
      title={Self-Distilled Reasoner: On-Policy Self-Distillation for Large Language Models}, 
      author={Siyan Zhao and Zhihui Xie and Mengchen Liu and Jing Huang and Guan Pang and Feiyu Chen and Aditya Grover},
      year={2026},
      eprint={2601.18734},
      archivePrefix={arXiv},
      primaryClass={cs.LG},
      url={https://arxiv.org/abs/2601.18734}, 
}

@misc{zhang2026guisd,
      title={Learn where to Click from Yourself: On-Policy Self-Distillation for GUI Grounding}, 
      author={Yan Zhang and Daiqing Wu and Huawen Shen and Can Ma and Yu Zhou},
      year={2026},
      eprint={2605.00642},
      archivePrefix={arXiv},
      primaryClass={cs.AI},
      url={https://arxiv.org/abs/2605.00642}, 
}

@misc{hendrycks2017baseline,
      title={A Baseline for Detecting Misclassified and Out-of-Distribution Examples in Neural Networks}, 
      author={Dan Hendrycks and Kevin Gimpel},
      year={2018},
      eprint={1610.02136},
      archivePrefix={arXiv},
      primaryClass={cs.NE},
      url={https://arxiv.org/abs/1610.02136}, 
}

@misc{guo2017calibration,
      title={On Calibration of Modern Neural Networks}, 
      author={Chuan Guo and Geoff Pleiss and Yu Sun and Kilian Q. Weinberger},
      year={2017},
      eprint={1706.04599},
      archivePrefix={arXiv},
      primaryClass={cs.LG},
      url={https://arxiv.org/abs/1706.04599}, 
}

@inproceedings{zeng2026fdcground,
  title={Fdc-ground: Improving grpo for gui grounding via exponential rewards and fact-aligned pruning},
  author={Zeng, Xiangjian and Li, Wenjing and Wu, Qingqiang and Zhang, Liang},
  booktitle={Proceedings of the AAAI Conference on Artificial Intelligence},
  volume={40},
  number={33},
  pages={28122--28130},
  year={2026}
}

@misc{zhao2025guicursor,
      title={Learning GUI Grounding with Spatial Reasoning from Visual Feedback}, 
      author={Yu Zhao and Wei-Ning Chen and Huseyin Atahan Inan and Samuel Kessler and Lu Wang and Lukas Wutschitz and Fangkai Yang and Chaoyun Zhang and Pasquale Minervini and Saravan Rajmohan and Robert Sim},
      year={2026},
      eprint={2509.21552},
      archivePrefix={arXiv},
      primaryClass={cs.CV},
      url={https://arxiv.org/abs/2509.21552}, 
}

@misc{zhu2026manyfaces,
      title={The Many Faces of On-Policy Distillation: Pitfalls, Mechanisms, and Fixes}, 
      author={Siqi Zhu and Xuyan Ye and Hongyu Lu and Weiye Shi and Ge Liu},
      year={2026},
      eprint={2605.11182},
      archivePrefix={arXiv},
      primaryClass={cs.AI},
      url={https://arxiv.org/abs/2605.11182}, 
}

@misc{ke2026respecting,
      title={Respecting Self-Uncertainty in On-Policy Self-Distillation for Efficient LLM Reasoning}, 
      author={Junlong Ke and Zichen Wen and Weijia Li and Conghui He and Linfeng Zhang},
      year={2026},
      eprint={2605.13255},
      archivePrefix={arXiv},
      primaryClass={cs.AI},
      url={https://arxiv.org/abs/2605.13255}, 
}

@misc{shi2026trustworthyguiagentssurvey,
      title={Towards Trustworthy GUI Agents: A Survey}, 
      author={Yucheng Shi and Wenhao Yu and Jingyuan Huang and Wenlin Yao and Wenhu Chen and Ninghao Liu},
      year={2026},
      eprint={2503.23434},
      archivePrefix={arXiv},
      primaryClass={cs.LG},
      url={https://arxiv.org/abs/2503.23434}, 
}

@misc{zheng2026scope,
      title={SCOPE: Signal-Calibrated On-Policy Distillation Enhancement with Dual-Path Adaptive Weighting}, 
      author={Binbin Zheng and Xing Ma and Yiheng Liang and Jingqing Ruan and Xiaoliang Fu and Kepeng Lin and Benchang Zhu and Ke Zeng and Xunliang Cai},
      year={2026},
      eprint={2604.10688},
      archivePrefix={arXiv},
      primaryClass={cs.LG},
      url={https://arxiv.org/abs/2604.10688}, 
}

@misc{yang2026selfdistilledrlvr,
      title={Self-Distilled RLVR}, 
      author={Chenxu Yang and Chuanyu Qin and Qingyi Si and Minghui Chen and Naibin Gu and Dingyu Yao and Zheng Lin and Weiping Wang and Jiaqi Wang and Nan Duan},
      year={2026},
      eprint={2604.03128},
      archivePrefix={arXiv},
      primaryClass={cs.LG},
      url={https://arxiv.org/abs/2604.03128}, 
}

@misc{yuan2026visionopd,
      title={Vision-OPD: Learning to See Fine Details for Multimodal LLMs via On-Policy Self-Distillation}, 
      author={Qianhao Yuan and Jie Lou and Xing Yu and Hongyu Lin and Le Sun and Xianpei Han and Yaojie Lu},
      year={2026},
      eprint={2605.18740},
      archivePrefix={arXiv},
      primaryClass={cs.CV},
      url={https://arxiv.org/abs/2605.18740}, 
}

@misc{tan2026paint,
      title={PAINT: Partial-Solution Adaptive Interpolated Training for Self-Distilled Reasoners}, 
      author={Zhiquan Tan and Yinrong Hong},
      year={2026},
      eprint={2604.26573},
      archivePrefix={arXiv},
      primaryClass={cs.LG},
      url={https://arxiv.org/abs/2604.26573}, 
}

@misc{bengio2015scheduled,
      title={Scheduled Sampling for Sequence Prediction with Recurrent Neural Networks}, 
      author={Samy Bengio and Oriol Vinyals and Navdeep Jaitly and Noam Shazeer},
      year={2015},
      eprint={1506.03099},
      archivePrefix={arXiv},
      primaryClass={cs.LG},
      url={https://arxiv.org/abs/1506.03099}, 
}

@misc{yang2026trontargetedruleverifiableonline,
      title={TRON: Targeted Rule-Verifiable Online Environments for Visual Reasoning RL}, 
      author={Tianze Yang and Yucheng Shi and Ruitong Sun and Jingyuan Huang and Ninghao Liu and Jin Sun},
      year={2026},
      eprint={2606.01599},
      archivePrefix={arXiv},
      primaryClass={cs.AI},
      url={https://arxiv.org/abs/2606.01599}, 
}

@misc{liu2025scalecua,
      title={ScaleCUA: Scaling Open-Source Computer Use Agents with Cross-Platform Data}, 
      author={Zhaoyang Liu and Jingjing Xie and Zichen Ding and Zehao Li and Bowen Yang and Zhenyu Wu and Xuehui Wang and Qiushi Sun and Shi Liu and Weiyun Wang and Shenglong Ye and Qingyun Li and Xuan Dong and Yue Yu and Chenyu Lu and YunXiang Mo and Yao Yan and Zeyue Tian and Xiao Zhang and Yuan Huang and Yiqian Liu and Weijie Su and Gen Luo and Xiangyu Yue and Biqing Qi and Kai Chen and Bowen Zhou and Yu Qiao and Qifeng Chen and Wenhai Wang},
      year={2025},
      eprint={2509.15221},
      archivePrefix={arXiv},
      primaryClass={cs.CV},
      url={https://arxiv.org/abs/2509.15221}, 
}

\clearpage
\appendix

\section{Prompt Templates, Privileged Visual Cues, and Training Targets}
\label{app:prompt_and_training_targets}

\subsection{Shared System Prompt}
\label{app:shared_system_prompt}

\begin{lstlisting}[style=promptstyle, caption={Shared system prompt used by both the teacher and the student.}, label={lst:shared_system_prompt}]
You may call one or more functions to assist with the user query.
You are provided with function signatures within <tools> ... </tools> XML tags:
<tools>
{"name": "computer_use", "description": "Use a mouse to interact with a computer.", "notes": "Click with the cursor tip centered on targets; avoid edges unless asked. Do not use other tools (type, key, scroll, left_click_drag). Only left_click are allowed.", "parameters": {"type": "object", "required": ["action"], "properties": {"action": {"type": "string", "enum": ["left_click"], "description": "The action to perform."}, "coordinate": {"type": "array", "description": "(x, y): pixels from left/top. Required for action=left_click."}}}}
</tools>

For each function call, return a JSON object with function name and arguments within <tool_call> ... </tool_call> XML tags:
<tool_call>
{"name": "<function-name>", "arguments": <args-json-object>}
</tool_call>
\end{lstlisting}

\subsection{Student and Teacher User Prompts}
\label{app:user_prompts}

\begin{lstlisting}[style=promptstyle, caption={Student user prompt template.}, label={lst:student_prompt}]
<image>
{original GUI instruction / query}
\end{lstlisting}

\begin{lstlisting}[style=promptstyle, caption={Teacher user prompt template with privileged hint.}, label={lst:teacher_prompt}]
<image>
{original GUI instruction / query} Hint: The answer is located within the green rectangle.
\end{lstlisting}

\begin{lstlisting}[style=promptstyle, caption={Example teacher user prompt.}, label={lst:teacher_prompt_example}]
<image>
5:20 PM Hint: The answer is located within the green rectangle.
\end{lstlisting}

\subsection{Training Target Format}
\label{app:target_format}

The model is trained to output a structured tool call. In our main experiments, the target response contains no additional natural-language reasoning or rationale. The canonical output format is:

\begin{lstlisting}[style=promptstyle, caption={Canonical training target format.}, label={lst:target_format}]
<tool_call>
{"name": "computer_use", "arguments": {"action": "left_click", "coordinate": [x, y]}}
</tool_call>
\end{lstlisting}

\section{Construction of Visual Privileged Information}
\label{app:visual_privileged_information}

\subsection{Teacher Visual Privileged Information}
\label{app:visual_privileged_construction}

Following GUI-SD~\citep{zhang2026guisd}, we provide the teacher with target-aware visual privileged information during training. Given the original GUI screenshot $I$ and the ground-truth target bounding box $b$, we construct a Gaussian soft mask around the target region. Let $d_b(u,v)$ denote the Euclidean distance from pixel $(u,v)$ to the bounding box $b$, where pixels inside the box have distance $0$. The masked image is computed as
\[
\begin{aligned}
    \alpha(u,v)
    &=
    \exp\left(
        -\frac{d_b(u,v)^2}{2\sigma^2}
    \right),\\
    I_{\mathrm{mask}}(u,v)
    &=
    \alpha(u,v) I(u,v).
\end{aligned}
\]
Here, $\sigma$ controls the spatial decay of the Gaussian mask. This operation keeps the target region fully visible while softly suppressing background regions farther away from the target.

In addition, we draw a green rectangle around the ground-truth target region and append a short textual hint indicating that the answer is located inside the rectangle. These visual and textual cues are used only for the teacher during training. The student always receives the original GUI screenshot and the original instruction, without any privileged visual cue.

\subsection{Visualization of Privileged and Non-Privileged Inputs}
\label{app:privileged_visualization}

\begin{figure}[t]
    \centering

    \begin{minipage}[t]{0.48\linewidth}
        \centering
        \includegraphics[width=\linewidth]{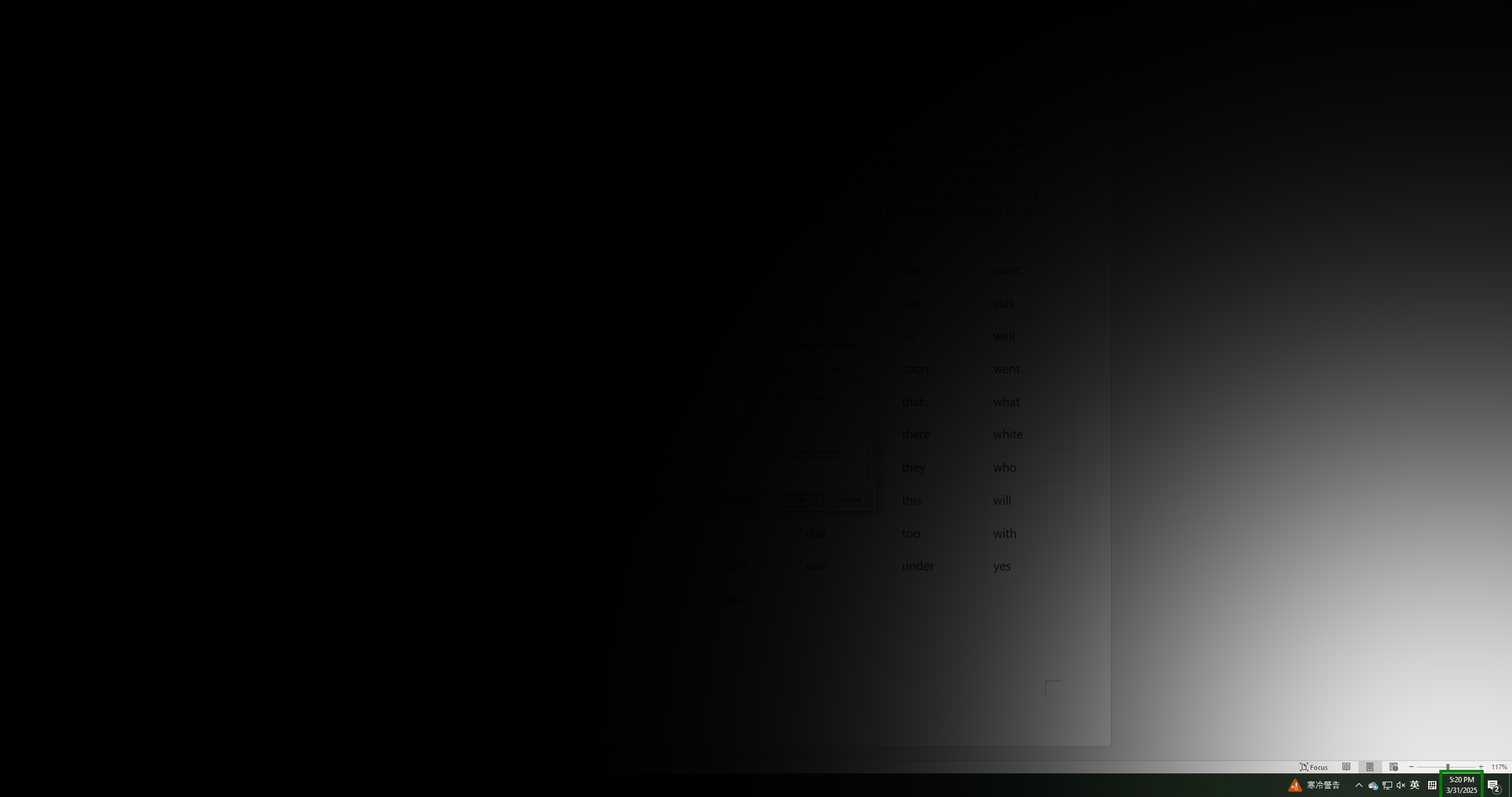}

        \vspace{0.4em}
        {\small (a) Teacher input with visual privileged information.}
    \end{minipage}
    \hfill
    \begin{minipage}[t]{0.48\linewidth}
        \centering
        \includegraphics[width=\linewidth]{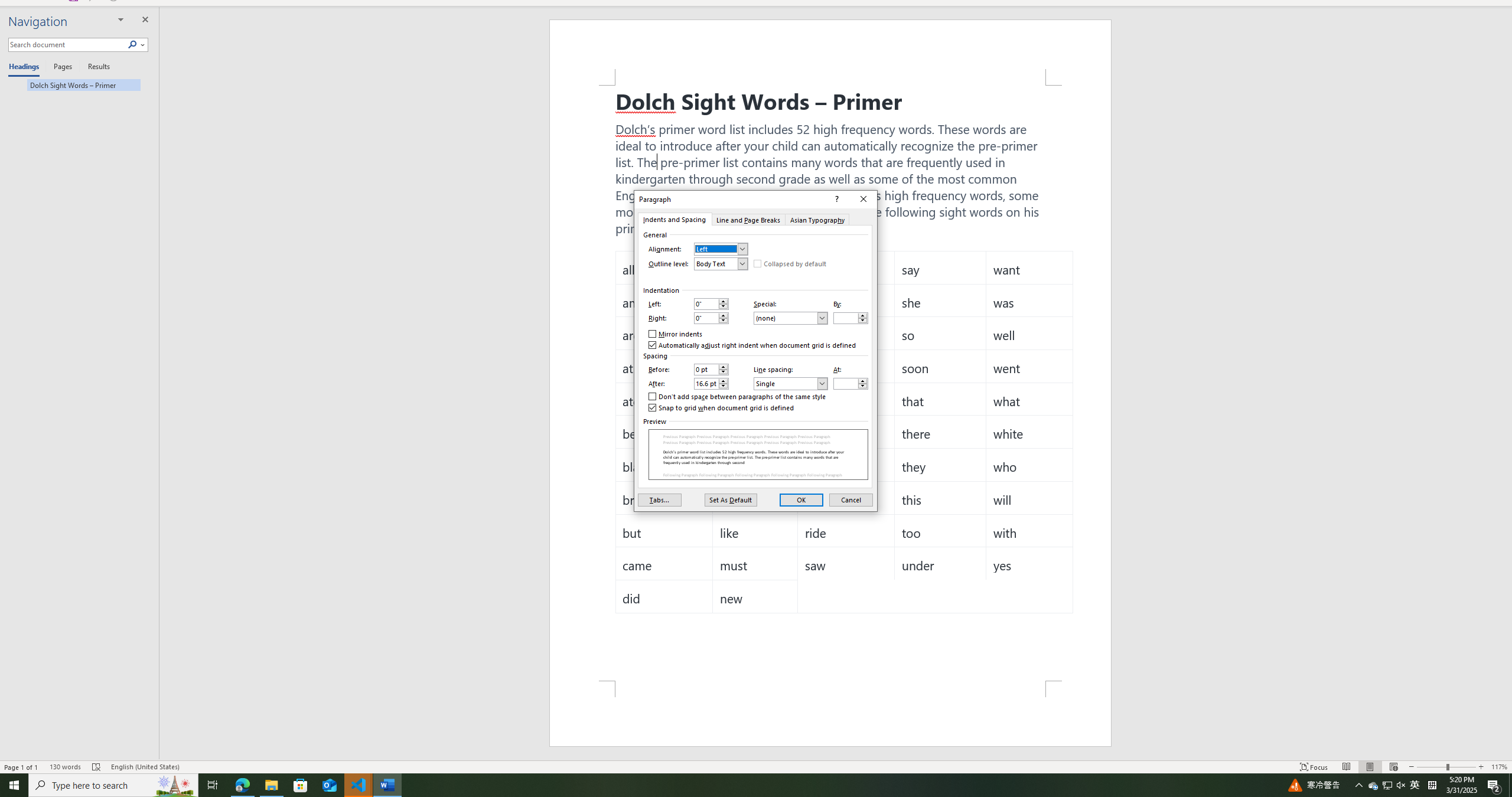}

        \vspace{0.4em}
        {\small (b) Student input without visual privileged information.}
    \end{minipage}

    \caption{
    Visualization of privileged and non-privileged inputs. 
    The teacher receives an augmented image in which the target region is marked by a green rectangle, together with a textual hint. 
    The student receives only the original GUI image and the original user instruction.
    }
    \label{fig:teacher_student_inputs}
\end{figure}

Figure~\ref{fig:teacher_student_inputs} shows an example of the teacher and student inputs. The teacher image contains the visual privileged information, while the student image remains the original GUI screenshot.

\section{Training Details}
\label{app:training_details}

\begin{table*}[!t]
\centering
\small
\begin{tabular}{ll}
\toprule
\textbf{Hyperparameter} & \textbf{Value} \\
\midrule
\multicolumn{2}{l}{\textit{Training setup}} \\
\midrule
Number of GPUs(H800) & 8 \\
Number of nodes & 1 \\
Total epochs & 1 \\
Final evaluated step & 62 \\
Train batch size & 112 \\
PPO mini-batch size & 112 \\
\midrule
\multicolumn{2}{l}{\textit{Optimization}} \\
\midrule
Optimizer & AdamW \\
Learning rate & $1.25 \times 10^{-6}$ \\
Learning-rate scheduler & Constant \\
Learning-rate warmup steps & 8 \\
Adam betas & $(0.9, 0.999)$ \\
Weight decay & 0.01 \\
Gradient clipping & 1.0 \\
\midrule
\multicolumn{2}{l}{\textit{Self-distillation}} \\
\midrule
Full-logit distillation & True \\
Distillation weight $\alpha$ & 1.0 \\
Distillation top-$k$ & None; full vocabulary \\
Distillation tail correction & True \\
Distillation clipping & None \\
Teacher source & Legacy model \\
Teacher always on & True \\
Teacher regularization & EMA \\
Teacher EMA decay & 0.95 \\
Student mixing coefficient & 0.05 \\
Teacher update rule &
$\theta_{\mathrm{teacher}} \leftarrow
0.95\,\theta_{\mathrm{teacher}} + 0.05\,\theta_{\mathrm{student}}$ \\
Teacher update frequency & After each successful optimizer step \\
Teacher prompt mode & None \\
Include environment feedback & False \\
Do not reprompt on self-success & True \\
Maximum reprompt length & 10,240 \\
Teacher entropy weighting & None \\
\bottomrule
\end{tabular}
\caption{Training hyperparameters for self-distillation.}
\label{tab:self_distillation_hparams}
\end{table*}

\begin{table*}[!t]
  \centering
  \small
  \setlength{\tabcolsep}{3pt}
  \begin{tabular}{lccccccc}
    \toprule
    GRPO step & SSP & ScreenSpot-v2 & UIEG & OSWorld-G & OSWorld-G-R & MMG & Avg \\
    \midrule
    150 & 65.46 & 92.92 & 31.36 & 63.48 & 71.45 & 83.59 & 68.04 \\
    240 & 65.53 & 93.55 & 32.83 & 63.65 & 70.92 & 83.31 & \textbf{68.30} \\
    330 & 65.40 & 93.32 & 32.43 & 62.94 & 69.68 & 84.03 & 67.97 \\
    436 (full epoch) & 65.02 & 93.24 & 32.65 & 62.06 & 70.39 & 83.87 & 67.87 \\
    \bottomrule
  \end{tabular}
  \caption{GRPO performance at four checkpoints during a full-epoch training run.}
  \label{tab:grpo_full_epoch}
\end{table*}

We provide the training hyperparameters used for our self-distillation experiments
in Table~\ref{tab:self_distillation_hparams}. We perform online self-distillation
using an EMA teacher. The teacher is initialized from a legacy model and updated
after each successful optimizer step according to
\[
\theta_{\mathrm{teacher}} \leftarrow
0.95\,\theta_{\mathrm{teacher}} + 0.05\,\theta_{\mathrm{student}}.
\]
Distillation is applied over the full vocabulary with weight $\alpha=1.0$.

\subsection{Baseline-Specific Training Details}

SFT and self-distillation methods are evaluated at their final checkpoints after one epoch of training, corresponding to 62 optimization steps under our setup. The main results report GRPO at step 62 for a comparison at the same number of optimization updates. For Naive-OPSD, the teacher is
provided with the ground-truth text bounding box as privileged information during
OPSD training. For GUI-SD, we directly adopt the training hyperparameters
reported in its original paper.

\paragraph{Full-epoch GRPO results.}
We additionally train GRPO for 436 steps on the same 6,990-example training dataset, providing approximately one epoch of prompt exposure. This run retains the original GRPO objective and main training hyperparameters, with the learning-rate schedule extended to the full training horizon. We evaluate checkpoints at steps 150, 240, 330, and 436 on all six benchmarks. As shown in Table~\ref{tab:grpo_full_epoch}, the best evaluated checkpoint, at step 240, achieves 68.30\% macro-average accuracy, while the final checkpoint achieves 67.87\%. Our method achieves 72.23\%, exceeding these results by 3.93 and 4.35 percentage points, respectively.

Our method's score aggregates five decoding runs, whereas each checkpoint of the extended GRPO run is evaluated once per benchmark. The runs have comparable prompt coverage, but differ in rollout counts and computational cost.

\end{document}